\documentclass[11pt,a4paper]{article}
\usepackage[utf8]{inputenc}
\usepackage[hyperref]{acl2020}
\usepackage{times}
\usepackage{latexsym}

\usepackage{microtype}
\usepackage{tabularx}
\usepackage{graphicx}
\graphicspath{ {./Images/} }

\aclfinalcopy 
\def\aclpaperid{119} 

\newcommand{\printfnsymbol}[1]{%
  \textsuperscript{{*}}%
}

\title{Exploring Pair-Wise NMT for Indian Languages}

\author{
    Kartheek Akella\thanks{*Equal Contribution. Sridhar worked on Tamil and Urdu, Himal worked on Gujarati, Kartheek worked on Malayalam and Marathi, Aman worked on Hindi and Punjabi and Zeeshan worked on Odiya. Himal was responsible for drafting this paper.} \\
  CVIT, IIIT-H \\
  \texttt{sukruthkartheek@gmail.com} \\\And
  Sai Himal Allu\printfnsymbol{1} \\
  CVIT, IIIT-H \\
  \texttt{saihimal.allu@gmail.com} \\\AND
  Sridhar Suresh Ragupathi\printfnsymbol{1} \\
  CVIT, IIIT-H \\
  \texttt{srsridhar.98@gmail.com} \\\And
  Aman Singhal \\
  CVIT, IIIT-H \\
  \texttt{amansinghalml@gmail.com}\\\AND
  Zeeshan Khan \\
  CVIT, IIIT-H \\ 
  \texttt{zeeshank606@gmail.com} \\\And
  Vinay P. Namboodiri \\
  University of Bath \\
  \texttt{vpn22@bath.ac.uk}\\\And
  C V Jawahar \\
  CVIT, IIIT-H \\
  \texttt{jawahar@iiit.ac.in}\\
}

\date{}

\begin{document}
\maketitle
\begin{abstract}
In this paper, we address the task of improving pair-wise machine translation for specific low resource Indian languages. Multilingual {\sc nmt} models have demonstrated a reasonable amount of effectiveness on resource-poor languages. In this work, we show that the performance of these models can be significantly improved upon by using back-translation through a filtered back-translation process and subsequent fine-tuning on the limited pair-wise language corpora. The analysis in this paper suggests that this method can significantly improve a multilingual models' performance over its baseline, yielding state-of-the-art results for various Indian languages.
\end{abstract}

\section{Introduction}
Neural machine translation ({\sc nmt}) algorithms, as is common for most deep learning techniques, work best with vast amounts of data. Various authors have argued that their performance would be limited for low resource languages 
\citet{ostling2017neural},
\citet{gu-etal-2018-universal}, \citet{kim-etal-2020-unsupervised}. One way to bridge this gap is through the use of multilingual {\sc nmt} algorithms to bypass the data limitations of individual language pairs (\citet{johnson-etal-2017-googles}, \citet{aharoni-etal-2019-massively}, \citet{vazquez-etal-2019-multilingual}). The use of such a model has been demonstrated recently by \citet{philip2020revisiting}. In this paper, we investigate the problem of improving pair-wise {\sc nmt} performance further over existing multilingual baselines. We specifically analyze the use of back-translation and fine-tuning to this effect. Our results suggest that it is possible to improve the performance of individual pairs of languages for various Indian languages. The performance of these language pairs is evaluated over standard datasets, and we observe consistent improvements over the two main baselines: an {\sc nmt} system trained pair-wise from scratch using the corpora available for the pair of languages, a multilingual {\sc nmt} model that uses many different languages.

\section{Previous Work}
The problem of multilingual {\sc nmt} has attracted significant research attention in the recent past. 
\citep{dong-etal-2015-multi} proposed the first multilingual model with a one-to-many mapping of languages, whereas \citep{ferreira-etal-2016-jointly} shared a single attention network for all language pairs. Recent works like \citep{lample2019cross}, \citep{conneau-etal-2020-unsupervised} which are an extension of \citep{roberta-corr-2019} and \citep{devlin2018bert} have improved upon the initial formulation of the multilingual {\sc nmt} problem. 
Works like \citep{currey-etal-2017-copied}, \citep{gen-nmt-nips2018}, \citep{li-eisner-2019-specializing} and \citep{hewitt-liang-2019-designing} use monolingual data to supplement their parallel corpora to build an {\sc nmt} system. 
On the flip side \citep{lample-2017-unsupervised}, \citep{wang-etal-2018-denoising},  \citep{artetxe2018iclr} study the unsupervised paradigm using only monolingual corpora.

Within the context of Indian languages, \citep{Chandola1994OrderedRF} and \citep{ILEHMT} were one of the first works to explore a rule-based approach for translation from Hindi to English whereas \citep{mt-ind-langs}, \citep{barman-etal-2014-assamese}, \citep{8464781} and \citep{choudhary-etal-2018-neural} have explored this problem through the prism of {\sc nmt}. 
\citep{philip2019baseline} and \citep{madaan-sadat-2020-multilingual} extend the concept of multilingual {\sc nmt} to the setting of Indian languages. 
Due to the recent efforts undertaken by the authors of \citet{kakwani2020indicnlpsuite}, Indian languages are now better represented in terms of available monolingual corpora. 
These resources set up a fertile ground for exploitation by semi-supervised and unsupervised {\sc nmt} approaches, which are consistent with the setting we study in this work.

\section{Method}

Consider a setting in which we have limited pair-wise corpora between a pair of languages, and we would like to obtain improved performance. We go about achieving this through the following procedure. 
First, we train a multilingual model on several languages. Next, we use existing monolingual corpora through back-translation ({\sc bt}), and then we fine-tune the model using available pair-wise corpora. 
We show that this particular procedure indeed improves over the alternative approach of training a pair-wise {\sc nmt} system using the available corpora. For the first step, we use the multilingual {\sc nmt} model provided publicly by \citep{philip2020revisiting}. We now provide details regarding the other two stages.

\subsection{Back Translation}
In {\sc nmt} literature, {\sc bt} is an effective approach that allows {\sc nmt} to pivot from a fully supervised setting to a semi-supervised setting. When supplemented with other objectives (autoencoder denoising \citep{artetxe2018iclr}, cross-translation \citep{Garca2020AMV}), {\sc bt} has demonstrated high efficacy in a fully unsupervised {\sc nmt} setting as well. In the semi-supervised paradigm, which we study in this work, our object of interest is to generate a meaningful learning signal from monolingual resources of a particular language that allows a reasonable {\sc nmt} model to exploit these resources to improve its performance on language pairs which include that specific language. 

\subsection*{Filtering Mechanism}
\begin{figure}[t]
\centering
\includegraphics[scale = 0.18]{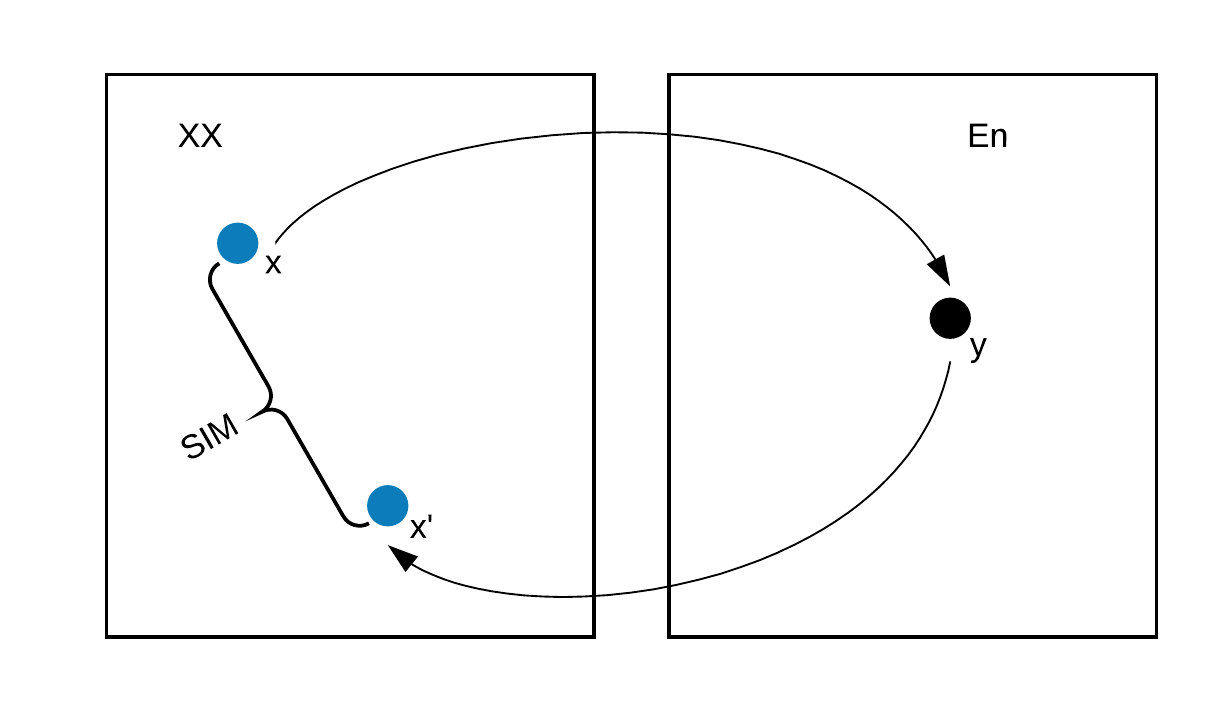}
\caption{SIM stands for a similarity heuristic. We use sentence wise {\sc bleu} scores in our work}
\label{fig:bt}
\end{figure}

Using a reasonable {\sc nmt} model, {\sc bt} can leverage monolingual resources to generate notable amounts of low-quality synthetic parallel data. If low-quality parallel corpora can be filtered through some means such that erroneous translation pairs are eliminated, we can obtain a strong learning signal from such a filtered corpus. To design such a filtering mechanism, we draw inspiration from generative modelling literature, precisely the idea of cyclic consistency \citep{Zhu2017UnpairedIT}. Briefly, the idea of cyclical consistency within the context of computer vision relates to minimizing the discrepancy between an image from domain X and the image obtained after transforming it to a domain Y and then converting it back to the domain X.  We adopt this approach to build our filtering mechanism, in which we first use our reasonable {\sc nmt} model to generate intermediate English ({\sc en}) translations for the sentences in the monolingual corpus of some language ({\sc xx}). As illustrated in fig \ref{fig:bt}, we then use these intermediate English translations to back-translate it into {\sc xx} and then evaluate the sentence wise {\sc bleu} \citep{papineni-etal-2002-bleu} scores for each such translation as a measure of similarity. Only sentences that cross an empirically chosen threshold are retained to ensure that the generated translations are of good quality to obtain a reasonably high-quality synthetic parallel dataset. We refer to this filtering scheme as {\sc xx-en-xx}. 

We initialise our {\sc nmt} model using the weights from the multilingual {\sc nmt} model provided by the authors of \citep{philip2020revisiting}. The authors train a Transformer \citep{vaswani2017attention} model on 10 Indian languages namely (notation in brackets), Hindi (hi), Telugu (te), Tamil (ta), Malayalam (ml), Urdu (ur), Bangla (bn), Gujarati (gu), Marathi (mr) and Odia (od) in addition to English (en).  They make two chief architectural decisions in this regard. One they develop a shared vocabulary over all languages of interest, giving equal representation to each language in the vocabulary (equal number of tokens from each language). The second, they share the encoder-decoder parameters of the Transformer model across all possible language pairs, a decision which encourages the model to learn a shared embedding space for all languages of interest. The low resource nature of these languages is primarily addressed through two techniques: namely, Transfer Learning and Backtranslation \citep{sennrich-etal-2016-improving}.
This design choice allows us to use the same {\sc  nmt} model for the {\sc xx-en} and the {\sc en-xx} directions of the {\sc xx-en-xx} setting with a consistent amount of effectiveness.  We reason that an {\sc en-xx-en} filtering scheme would result in a compounding of errors problem due to the superior performances offered by the multilingual {\sc nmt} model in the {\sc xx-en} direction in contrast to the {\sc en-xx} direction. In such a case, populating our filtered corpus would require selecting a lower value of the threshold, which would compromise the quality of translation pairs, thereby leading to a weak supervisory signal.

\begin{table}
\centering
\begin{tabular}{ccc}
\hline \textbf{ }  & \small \textbf{Pre-filt \#pairs} & \small \textbf{Post-filt \#pairs} \\ \hline
\small hi & \small 4M & \small 140K \\
\small pa & \small 58K & \small 7K \\
\small mr & \small 178K & \small 58K \\
\small gu & \small 370K & \small 39K \\
\small ta & \small 88K & \small 34K \\
\small ur & \small 400K & \small 105K \\
\small ml & \small 178K & \small 52K \\
\small od & \small 221K & \small 64K  \\
\hline
\end{tabular}
\caption{\label{table:t1} Monolingual corpora utilized}
\end{table}

\setlength{\tabcolsep}{3pt}

\begin{table}
\centering
\begin{tabularx}{\columnwidth}{llXXllXXX}
\hline \textbf{ }  & \small \textbf{hi} & \small \textbf{pa} &  \small \textbf{gu} & \small \textbf{mr} & \small \textbf{ta} & \small \textbf{ml} & \small \textbf{ur} & \small \textbf{od} \\
\hline
\small iitb & \small 1.5M & \small -  & \small - & \small - & \small - & \small - & \small-  & -\\
\small cvit-pib & \small 195K & \small 27K & \small 29K & \small 81K & \small 87K & \small 32K & \small 45K & -\\
\small ufal & - & - & - & - & \small 167K & - & - & - \\
\small ilci & \small 49K &\small 49K  & \small 49K & - & \small 49K & \small 30K & \small 49K & -\\
\small odcorp1.0 & - & - & - & - & - & - & - & \small 27K \\
\small odcorp2.0 & - & - & - & - & - & - & - & \small 97K \\
\hline
   & \small 1.75M & \small 76K & \small 79K & \small 81K & \small 303K & \small 62K & \small 94K & \small 124K \\
\hline
\end{tabularx}
\caption{\label{table:t2} Parallel corpora utilized}
\end{table}

\section{Experimental Setup}
\label{sec:expreiments}

\subsection{Training Details}
We make use of the Transformer-Base, a part of Fairseq library \citep{ott2019fairseq}, which is built with 6 encoder-decoder layers, each having 512 hidden units and a singular attention head as our {\sc nmt} model. We initialise our model with the weights of the multilingual {\sc nmt} model provided by the authors of \citep{philip2020revisiting}. We also utilise the SentencePiece \citep{Kudo2018SentencePieceAS} models of \citep{philip2020revisiting} to build our vocabulary. 

For all languages of interest, we carry out filtering of the back-translated corpus by first evaluating the mean of sentence-wise {\sc bleu} scores for the cyclically generated translations and then selecting a value slightly higher than the mean as our threshold. Sentences that cross this threshold are then included along with their corresponding translations in our filtered corpus. We supplement the training of our {\sc nmt} model on a filtered back-translated corpus with two rounds of finetuning on a relevant parallel corpus: a pre-training phase and a post-training phase before carrying out the final evaluation. We reason that a pre-training step enhances the possibility of generating a high-quality synthetic filtered corpus from the related monolingual corpora by providing a more robust prior {\sc nmt} model for the {\sc bt} routine.  A post-training step ensures that the {\sc nmt} model is subjected to a more reliable supervisory signal before the final evaluation is carried out. We train all our models using AdamW \citep{loshchilov2019decoupled} optimizer until a local minimum is achieved.
\begin{table*}
\centering
\begin{tabular}{c|c|llll|l|l|l}
\hline
&&\multicolumn{4}{c}{\small State of the Art} & \multicolumn{3}{c}{\small {\sc nmt} (Different  Attempts)}\\
\hline
\tiny PAIRS  & \tiny Test-Set & \multicolumn{4}{c}{\small Top-4 (Prev. Attempts)} & \tiny Rand-init & \tiny {\sc m-nmt}$^1$ & \tiny Filt-BT \\ \hline
\tiny En-Hi & \small {\sc mkb} & \small 15.65$^2$ & \small 16.23$^2$ & \small 21.05$^2$ & \small \textbf{24.48}$^2$ & \small 13.28 & \small 16.93 &  \small 16.67 \\ \hline
\tiny En-Pa & \small {\sc ilci} & &&& \small 23.05$^1$ &\small 10.67 & \small 21.36 &  \small \textbf{23.52} \\ \hline
 \tiny En-Mr & \small {\sc mkb} &  \small 8.79$^2$ & \small 8.84$^2$ & \small 8.97$^2$ & \small 9.65$^2$ &  \small 2.77 & \small 9.84 &  \small \textbf{9.89} \\ \hline
\tiny En-Gu
 & \small {\sc mkb} & \small 9.73$^2$ & \small 10.13$^2$ & \small 11.24$^2$ & \small 11.70$^2$ & \small 2.63 & \small 12.92 & \small \textbf{14.37} \\ \hline
\tiny En-Ta & \small {\sc mkb}  & \small 4.33$^2$& \small 4.43$^2$ & \small 4.53$^2$ & \small 4.94$^2$ & \small 0.78 & \small 4.86 & \small \textbf{5.69} \\ \hline
\tiny En-Ta & \small {\sc ufal}  & \small 11.73$^2$& \small 12.51$^2$ & \small 12.74$^2$ & \small 13.05$^2$ & \small 0.78 & \small 7.80 & \small \textbf{19.07} \\ \hline
\tiny En-Ml & \small {\sc mkb} &  \small 5.00$^2$ & \small 5.17$^2$& \small 5.42$^2$ & \small 6.32$^2$ & \small 1.59 & \small 2.65 &  \small \textbf{6.40} \\ \hline
\tiny En-Ur & \small {\sc mkb} &&&& \small 22.16$^1$ &\small 3.90 & \small 22.16 &  \small \textbf{24.76} \\ \hline
\tiny En-Od & \small {\sc odiencorp}\tiny v2 & \small 7.93$^2$ & \small 9.35$^2$ & \small 9.85$^2$ & \small \textbf{11.07}$^2$ & \small 5.29 & \small 0.96 & \small 10.84 \\
\hline
\end{tabular}
\caption{ Comparison of our {\sc nmt} results with others publicly available on {\sc wat} leader board $^2$. For results that were not available on {\sc wat} leaderboard (Pa,Ur), we compare it with results from the paper \citep {philip2020revisiting}. We find that initialisation using a multlilingual model$^1$ is highly effective for {\sc nmt} in contrast to initialising randomly and training only on the respective language}
\label{table:t3}
\end{table*}
\subsection{Datasets}
For Hi, Od, and Ta, we use the {\sc iit-b} corpus, {\sc OdiEnCorp-v1.0}, and PMIndia Corpus \citep{haddow2020pmindia} respectively. 
For the rest of the languages, we utilise the relevant monolingual corpora provided by the authors of \citep{kakwani2020indicnlpsuite}.
We use only a part of these monolingual corpora in our experiments, the statistics of which we present in Table \ref{table:t1}. 
Our pre-training and post-training routines dictate the need for a parallel corpus for all our languages of interest. 
Due to the relative lack of availability of large high-quality parallel corpora for our language pairs of interest, we collate available resources to obtain a final parallel corpus, the details of which are presented in Table \ref{table:t2}. 
In addition to {\sc cvit-pib} \citep {siripragada-etal-2020-multilingual} and {\sc ilci} \citep {jha-2010-tdil} datasets, we also utilise {\sc iit-b} Hi-En corpus \citep {kunchukuttan-etal-2018-iit} for Hi, {\sc ufal EnTamv2.0} \citep {biblio:RaBoMorphologicalProcessing2012} for Ta and {\sc OdiEnCorp} 1.0 and 2.0 \citep {parida-etal-2020-odiencorp} for Od. 
For evaluation we use the {\sc cvit-mkb} \citep {siripragada-etal-2020-multilingual} dataset for the languages in {\sc mkb}. 
We evaluate on the {\sc ilci} dataset for Pa, OdiEncorp-v2.0 \citep {parida-etal-2020-odiencorp} for Od and {\sc ufal} EnTamv2.0 \citep {biblio:RaBoMorphologicalProcessing2012} for Tamil.

\section{Results and Discussions}

We report {\sc bleu} scores on all the test sets specified. 
We refer to our approach as Filt-BT in Table \ref{table:t3} and contrast our results with a randomly initialised model trained from scratch with the same conditions (Rand-Init), the multilingual model that we use as our prior {\sc nmt} model ({\sc m-nmt}) 
\footnote{\citet{philip2020revisiting}} and the top 4 publicly available results on the {\sc wat} leaderboard
\footnote{\url{http://lotus.kuee.kyoto-u.ac.jp/WAT/WAT2020/}}
\footnote{We do not include results from model-ensembling approaches}. 
Since Pa and Ur do not have an entry on the leaderboard, we instead make a comparison with the results reported in  \citep{philip2020revisiting}, which are the present {\sc sota} results to the best of our knowledge.

The first comparison highlights the benefits of warm-starting our {\sc nmt} model from a {\sc m-nmt} model, whereas the second comparison helps us ascertain the efficacy of filtered {\sc bt} and as such, we report consistent gains over both these baselines for all the language pairs. 
In all the language pairs barring Odia, we demonstrate the superior performances of a prior multilingual model in contrast to a specialized model trained from scratch, validating our claim that initialization using a multilingual model is highly effective for {\sc nmt} in contrast to initializing randomly and training only on the respective language. 
Typically, we observe that using high threshold values for filtering leads to the filtered corpus getting biased by selecting comparatively shorter sentences. 
To maintain a healthy mix of both types of sentences, we use a threshold value slightly higher than the mean of the sentence-wise {\sc bleu} scores which we find in our experiments empirically provides for a more balanced high quality (in terms of translation quality) corpus thereby guaranteeing a better supervisory signal.

For Ta and Ur, we notice a massive boosts in performance (11.88 and 10.49 {\sc bleu} points respectively) over our multilingual baseline, with significant gains, also being noticed for Ml and Gu. The {\sc m-nmt} model, which we use to initialize our {\sc nmt} model, has been trained on OdiEnCorpv1.0, whereas our Rand-Init model has been trained using both versions of the dataset. We ascribe the former's inferior performance compared to the latter on Odia to the domain mismatch between both these versions, something which the latter model does not have to face.
We select a subset of the monolingual data to maintain consistency with our computing resources. For 200K sentences, we train on a 1080ti {\sc nvidia gpu} and found that back translation took about 3 hours. We decided that it would be an adequate sample size to test the validity of our approach. Since only a subset of monolingual data provided by \citep {kakwani2020indicnlpsuite} is used, we fully expect these results to trend upwards if the entire corpora were to be utilised. We indicate the {\sc sota} performance for each language in bold. As such, we achieve {\sc sota} performances on Pa, Gu, Ml, Mr, Ta and Ur. 

\section{Summary and Directions}
Our explorations in the applicability of Neural Machine Translation for Indian languages lead to the following observations (i) Multilingual models are a promising direction to address data scarcity and the variability of resources across languages (ii) adapting a multilingual model for a specific pair can provide superior performances. We believe these solutions can further benefit from the availability of mono-lingual resources and noisy parallel corpora. 

\bibliography{anthology,acl2020}
\bibliographystyle{acl_natbib}

\end{document}